\documentclass[letterpaper, 10 pt, conference]{ieeeconf}  

\IEEEoverridecommandlockouts                              

\usepackage{amsmath,amssymb,amsfonts}
\usepackage{xcolor}
\usepackage{url}
\usepackage{siunitx}
\usepackage{graphicx}
\usepackage{dblfloatfix}
\usepackage{float}
\graphicspath{ {./figures/} }
\usepackage{cite}
\usepackage{subfigure}

\usepackage{textcomp}

\usepackage{hyperref}

\usepackage{bm}
\usepackage{float}
\usepackage[font=small,skip=7pt]{caption}
\usepackage[linesnumbered,noend,ruled,vlined]{algorithm2e}
\usepackage{microtype}
\usepackage{multirow}
\usepackage{graphicx}
\usepackage{rotating}
\SetKwInput{KwInput}{Input}                
\SetKwInput{KwOutput}{Output}              
\newlength\fwidth
\usepackage{mathtools}

\usepackage{siunitx}
\usepackage{epstopdf}
\usepackage{mathtools}
\usepackage[T1]{fontenc}
\DeclareGraphicsExtensions{.pdf,.eps}
\usepackage{textcomp}
\usepackage[ISO]{diffcoeff}
\usepackage{caption}
\usepackage{amsmath}

\def\BibTeX{{\rm B\kern-.05em{\sc i\kern-.025em b}\kern-.08em
    T\kern-.1667em\lower.7ex\hbox{E}\kern-.125emX}}

\usepackage[acronym,nomain,nonumberlist]{glossaries}
\usepackage{url}

\title{\LARGE \bf

STAGE: Scalable and Traversability-Aware Graph based Exploration Planner for Dynamically Varying Environments}

\author{Akash Patel, Mario A.V. Saucedo, Christoforos Kanellakis, and George Nikolakopoulos
\thanks{The authors are with Robotics \& AI Team, Department of Computer, Electrical and Space Engineering, Lule\r{a} University of Technology, Lule\r{a} SE-97187, Sweden. 
        {Corresponding author: \tt\small akapat@ltu.se}}%
}

\begin{document}

\maketitle
\thispagestyle{empty}
\pagestyle{empty}

\begin{abstract}


In this article, we propose a novel navigation framework that leverages a two layered graph representation of the environment for efficient large-scale exploration, while it integrates a novel uncertainty awareness scheme to handle dynamic scene changes in previously explored areas. The framework is structured around a novel goal oriented graph representation, that consists of, i) the local sub-graph and ii) the global graph layer respectively. The local sub-graphs encode local volumetric gain locations as frontiers, based on the direct pointcloud visibility, allowing fast graph building and path planning. Additionally, the global graph is build in an efficient way, using node-edge information exchange only on overlapping regions of sequential sub-graphs. Different from the state-of-the-art graph based exploration methods, the proposed approach efficiently re-uses sub-graphs built in previous iterations to construct the global navigation layer. Another merit of the proposed scheme is the ability to handle scene changes (e.g. blocked pathways), adaptively updating the obstructed part of the global graph from traversable to not-traversable. This operation involved oriented sample space of a path segment in the global graph layer, while removing the respective edges from connected nodes of the global graph in cases of obstructions. As such, the exploration behavior is directing the robot to follow another route in the global re-positioning phase through path-way updates in the global graph. Finally, we showcase the performance of the method both in simulation runs as well as deployed in real-world scene involving a legged robot carrying camera and lidar sensor.

\end{abstract}


\section{Introduction}\label{sec:introduction}

While the exploration capabilities of autonomous robots have been increasingly improving in recent years through novel navigation approaches, the design of exploration missions is underutilized in terms of environment representation. The majority of the exploration algorithms are developed with the intent to increase exploration metrics by remodeling exploration goal and/or path selection process. Traditional frontiers~\cite{yamauchi1997frontier} or sampling~\cite{lindqvist2021exploration} based exploration approaches have proven to perform well in small scale, less complex environments. However, in subterranean environments such as underground corridors, tunnels, mines and caves, there exist additional factors such as degraded perception, complex topology, multi-branched self-similar structures, dynamic obstacles that challenge autonomous exploration and mapping operations of robots. The DARPA organized Subterranean Challenge \cite{agha2021nebula}, \cite{rouvcek2020darpa}, \cite{tranzatto2022cerberus}, \cite{hudson2021heterogeneous} aimed to contribute in advancing the capabilities of autonomous robots in real life underground environments for Search And Rescue (SAR) missions. 
Autonomous deployment of robots in scenarios such as SAR missions, require the navigation autonomy to consider efficient exploration along with appropriate representation of the acquired knowledge of the explored environment. The dynamic change in the SubT environment also seen in the DARPA Subterranean challenge was a major obstruction for autonomous robots repositioning to a frontier in exploration. As part of development efforts and lessons-learnt from DARPA SubT challenge of team COSTAR, we propose the \textbf{S}calable and \textbf{T}raversability-\textbf{A}ware \textbf{G}raph based \textbf{E}xploration (STAGE) planner, that targets exploration proposing a memory efficient mechanism  for fast local sub-graphs generation and minimal information exchange (collision checks on overlapping sub-graphs) to update the global graph. The merit of this representation of the environment allows for dynamical updates of the global graph in situations of scene changes (e.g. blocked path). As such, to the best of the authors knowledge STAGE enables long-term autonomous robot navigation in GPS-denied and dynamically varying environments. 


\section{Related Works}

Building upon the original idea of frontiers in robotics context, several algorithms \cite{holz2010evaluating} present different exploration approaches by either selecting a candidate frontier or selecting a path to candidate frontier to demonstrate efficient exploration using frontiers based method. Apart from frontiers, there exist also sampling based approached where randomly sampled nodes are connected to form a tree structure spanning from robot's current position as root. Such approaches are generally classified as variant of Rapidly exploring Random Trees or RRT \cite{umari2017autonomous} that consider associated information gain with the nodes to compute local exploration goal. Both frontiers and sampling based approaches have their advantages and disadvantages based on the chosen environment. The frontiers based approach focuses on rapid exploration of unknown environments \cite{patel2023ref}, \cite{patel2023towards} but suffer from partially exploring small areas in the map. On the other hand, the sampling based \cite{bircher2016receding} approaches tend to efficiently cover local areas while overall performance is degraded on large scale environments. There exist also hybrid approaches\cite{lu2022optimal} that combine both frontiers and Next Best View legacy to form an exploration approach that rapidly explores small scale unknown areas. 

As part of development efforts within the DARPA SubT challenge, the competing teams demonstrated legged and aerial robot exploration through approaches tailored to fit subterranean settings. Most teams utilized local and global exploration planners where local planner drives the exploration through frontiers and/or RRT by computing A* paths to goals and utilize full explored map to plan computation heavy global or homing path in case of dear end or exceeding robot endurance limits \cite{roucek2021system}. The authors in \cite{dang2019graph} [GB planner] proposed a novel graph based exploration planner that constructs a graph through random sampling in adaptive oriented bounding box for local RRT-style exploration while building a global graph from previously visited nodes to plan global re-positioning and homing paths. The experimental evaluation of approach in \cite{dang2020graph} has proven to be the state-of-the-art in terms of adaptive exploration in unknown SubT environments. However, the GB planner still has limitations in terms of A) increasing path computation time on large scale exploration missions due to single layer navigation graph and B) Unable to adapt to change in the environment where global graph is already built. Through the capabilities of the proposed multi layer graph representation, traversability factor based adaptive path improvement and efficient planning, the STAGE planner aims to address open challenges in autonomous navigation in SubT environments.


\section{Contributions}

Based on the aforementioned state-of-the-art works in the field of subterranean exploration, the key contributions and highlights of the STAGE planner are mentioned below. 

The first contribution stems from the novel goal oriented graph generation that distinguishes STAGE planner from classical sampling based exploration approaches. The STAGE planner utilizes direct point cloud visibility to extract constantly updating frontier nodes to adaptively tune the local bounds for rapid graph building and graph improvement. Based on the frontier goals, an adaptively tuned sub-graph is generated for navigating the robot towards local volumetric gain-rich area. Furthermore, the proposed graph navigation scheme \textit{re-utilizes} previously built sub-graphs to save additional computation at global graph building stage. The STAGE planner extracts overlapping regions from previously built sub-graphs and performs collision checks \textit{only on the overlapping regions} to stitch the graphs together forming a dense global graph. 


The second contribution of the STAGE planner is derived to address the particular challenge of how the robot can adapt to unknown change in the global map while re-visiting previously visited places in the global re-positioning or homing manoeuvres. Nearly all methods in literature adapt either back tracking on previously visited positions or utilizing a global graph for navigation to global re-positioning frontier. In real life SAR missions, the environment may change during the mission and the global graph built in previously seen as \textit{Traversable} zone may be in collision with obstacle or occupied zones. To address this, we propose iterative update on execution path built from path ways of global graph. In contrast to blindly following the global path, the STAGE planner segments the global path into smaller segments with size proportional to the current point cloud visibility. Reusing the vertices of the segment of the path and sampling sparse vertices around the segment to assure a traversable path from robot current state to the end of the current segment. If such segment is found to be in untraversable zone due to the unknown change in the environment, the planner discards the current path and re-computes global path ways by taking into account the change in the global map.



\section{STAGE planner architecture} \label{sec:methodology}


\begin{figure*}[t]
    \centering
        \includegraphics[width=\textwidth]{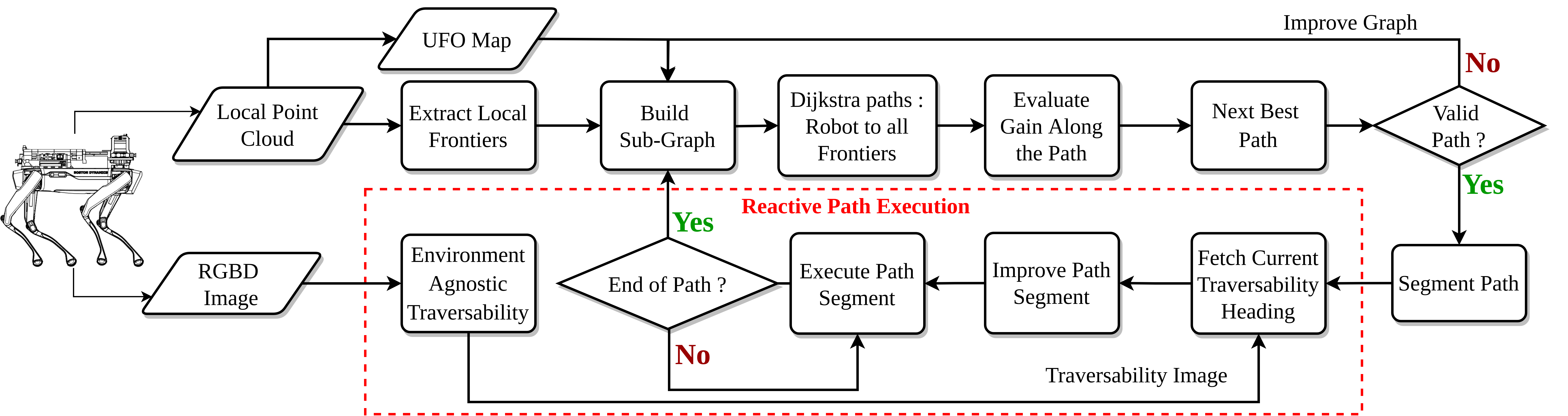}
    \caption{Overview of the STAGE Local Exploration Planning scheme.}
    \label{fig:local_glow_chart}
    \vspace{-1.5em}
\end{figure*}

The exploration of unknown subterranean environments is a problem broadly studied during the DARPA SubT challenge enabling the robotized autonomy for exploring and mapping the unknown underground terrains. The two major autonomy challenges observed during the competition were, A) Efficient exploration-path planning in varying topology of SubT settings, B) Global re-positioning or homing path planning in presence of unknown change in the environment. 
Let us denote the current point cloud scan as $\mathcal{P}$, the local 3D occupancy map as $\mathbb{M_{L}}$, explored global 3D map as $\mathbb{M_{G}}$ and set of potential frontier nodes as $\mathcal{F}$. where, $\mathcal{P} = \{ (x_{i}, y_{i}, z_{i})\ |\ i \in N \}$. The occupancy maps are often constructed with octree data structure where, a voxel $v$ with three possible states $v_{f}$, $v_{o}$ and $v_{u}$ (corresponding to occupancy probability) is subdivided into eight voxels until a minimum volume defined as the resolution of the octree $v_{res}$ is reached. From the current point cloud $\mathcal{P}$ and local occupancy map $\mathbb{M_{L}}$, $(\mathbb{M_{L}} \subset \mathbb{M_{G}})$ the set of frontiers $\mathcal{F}$ with at least $k$ number of unknown neighbouring voxels need to be extracted and inserted into $\mathbb{M_{G}}$. Based on the unknown neighbours density of a frontier voxel, the volumetric exploration gain $V_{gain}$ is computed. In order to perform rapid local exploration, the current visibility of the robot surroundings is depicted through local point cloud scan to orient the exploration direction in volumetric gain rich area. The navigation from the robot's current state $\zeta_{i} = \{(x_{i}, y_{i}, z_{i}, \psi_{i})\ |\ i \in N \}$ to a potential frontier, it builds a traversable sub-graph $\mathbb{S_{G}}^{i}$ composed of nodes $\mathcal{N} = \{ n_{1}, n_{2}, .. , n_{n} \}$ and edges $\mathcal{E} = \{ e_{1}, e_{2}, .. , e_{n} \}$. A local exploration path $\lambda_{L}$ to a potential frontier in field of view is chosen among the local path ways $P_{L}$ to all local frontiers. Throughout the paper, $var_{i}$ denotes the value of variable $var$ at $i^{th}$ planning iteration. 
\begin{figure*}[b]
    \centering
        \includegraphics[width=\textwidth]{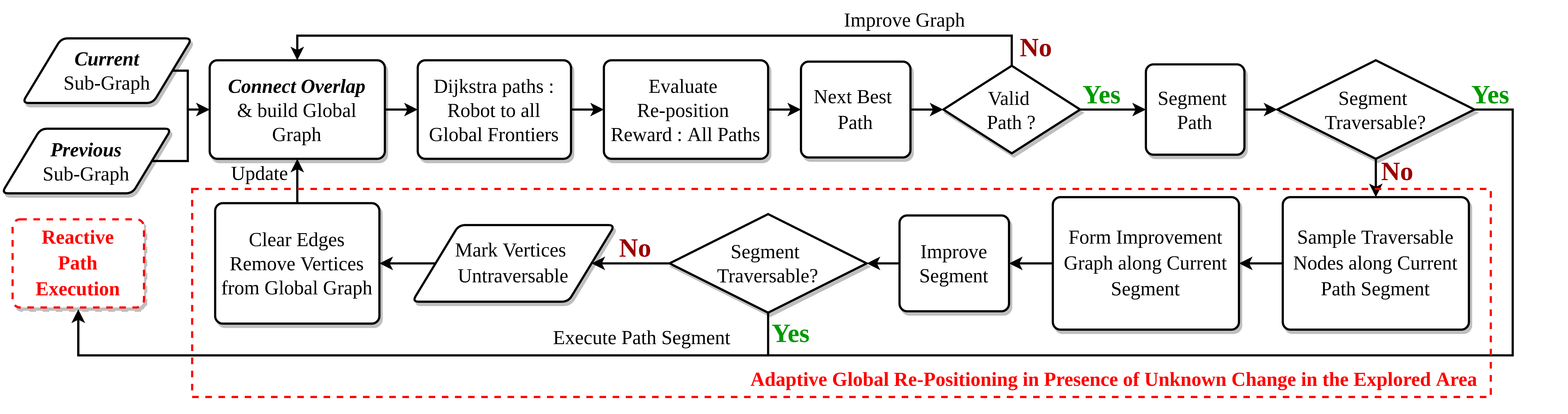}
    \caption{Overview of the STAGE Global Re-positioning Planning scheme.}
    \label{fig:local_glow_chart}
    \vspace{-1.5em}
\end{figure*}
Given the tube like multi branched topology of the subterranean settings, it is often observed that robots reach local dead ends in exploration, requiring efficient global re-positioning at a potential frontier in the explored map to continue the mission. For global navigation in known map, a dense global graph $\mathbb{G}_{G}$ is built to find multiple paths to multiple global frontiers as well as the location of the mission base station. Subject to the robot's latest knowledge of the global occupancy map $\mathbb{M_{G}}$, the robot queries path-ways to all potential global-repositioning candidates $\mathcal{C}$ where, $\mathcal{C} \subset \mathcal{F}$. Based on the volumetric gain evaluation of the re-positioning target frontier candidate and the information gain $I$ along the path, a global re-positioning path-way $\lambda_{G}$ needs to be computed to assure continued exploration. In contrast to blindly following the path $\lambda_{G}$, the important aspect of adaptive global re-positioning is to continuously assess the uncertainties $\Upsilon_{u}$ while navigating along pre-computed $\lambda_{G}$ to assure collision free path execution step.
\subsection{Local Exploration} \label{subsec:local_exploration}
\begin{figure}[t]
    \centering
        \includegraphics[width=\linewidth]{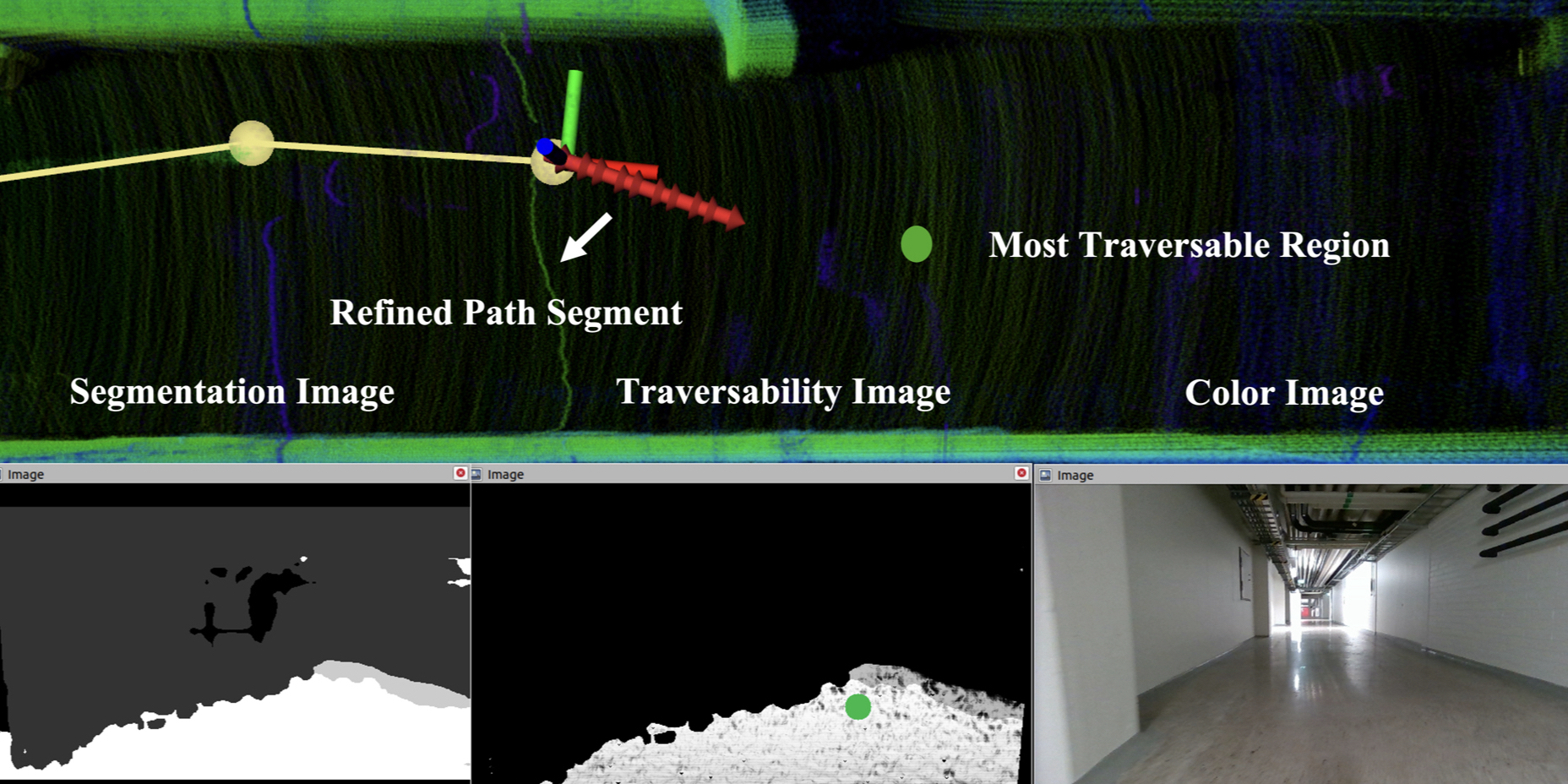}
    \caption{Semantic Traversability $\mathbb{ST}$ image. The refined path segment (red arrow) is computed based on the most traversable zone depicted on the $\mathbb{ST}_{image}$, where traversability is represented by a set of semantic labeled classes describing the smoothness of the terrain (e.g. flat, rough, obstacle, etc.).}
    \label{fig:traversibility_image}
    \vspace{-1.5em}
\end{figure}

At depicted in Fig. \ref{fig:local_glow_chart}, the inputs to the STAGE local exploration stage are local point cloud scan $\mathcal{P}$ and robot current state $\zeta$. Local map $\mathbb{M_{L}}$ update iterates through only the $\mathbb{M_{L}}$ to fetch new frontier voxels. Based on $\mathcal{P}$, a direct Pointcloud visibility based Planning space ($P_{v}P_{s}$) is tuned to iterate through local UFOMap~\cite{duberg2020ufomap} in order to sample free nodes $\mathcal{N}$ surrounding the robot. Taking into consideration the limited locomotion of legged robot, the nodes are only sampled in 2.5D allowing small elevation change from $\zeta_{z}$ for legged robot and in 3D for an aerial robot. Further considering the type of robot as input parameter, the planner forms connections between $n$ and $n_{neighbour}$ where node $n \in \mathcal{N}$. A sub-graph $\mathbb{S_{G}}$ for local navigation is built with $\mathcal{N}$ and $\mathcal{E}$. In order to reduce computational overhead, the nodes in $\mathcal{N}$ are examined with $kdtree$ data structure to find if a node has minimum required neighbouring voxels that are frontiers. Using the Dijkstra shortest path \cite{chen2003dijkstra}, shortest paths to all frontier nodes in sub-graph are computed. 
\begin{equation}
\small
\label{eqn:local_reward} 
\begin{split}
    \mathbb{R} =\overbrace{ W_{i} * log(\frac{e^{W_{u} * v_{u}} + e^{W_{f} * v_{f}}} { e^{W_{o} * v_{o}}})}^\text{Volumetric gain} - \\
    \overbrace{ W_{h} * (tan^{-1}(\frac{\zeta_{y} - n_{y}}{\zeta_{x} - n_{x}}) - \zeta_{\psi}) }^\text{Heading penalty}
\end{split}
\end{equation}
\normalsize
To further assure local rapid exploration, we constrain the robot's field view to narrow down search for informative frontier nodes. A local exploration reward $\mathbb{R}$ is computed through (\ref{eqn:local_reward}) that takes into account the volumetric gain associated with the node in sub-graph and deviation from the robot's current heading direction $\zeta_{\psi}$. In (\ref{eqn:local_reward}), $W_{i}, W_{u}, W_{f}, W_{o}$ and $W_{h}$ are the weights associated with information gain, unknown, free, occupied and heading deviation. Among the local path ways $P_{L}$ a local exploration path $\lambda_{L}$ to a node with highest local exploration reward $\mathbb{R}$ is computed. 
Since the planning is done through collision checks based on ray casting, STAGE planner introduces a reactive path execution step to assure robot-safe path. The path $\lambda_{L}$ is further segmented and the current segment in the path $\lambda_{L}^{k}$ is refined to adapt the traversability information. The end point of path segment $\lambda_{L}^{k}$ is shifted to the most traversable pixel in the Semantic Traversability image ($\mathbb{ST}_{image}$) as shown in Fig. \ref{fig:traversibility_image}. Using the local exploration algorithm proposed in \autoref{algo:local}, in an iterative manner local exploration is continued until there is no new local information gain associated to frontiers in field of view. At this moment, a global re-positioning path is requested from global re-positioning planner to relocate the robot to a potential frontier of $\mathbb{M_{G}}$.  

\DontPrintSemicolon
\begin{algorithm}
\small
\SetAlgoLined
\caption{Local Exploration} \label{algo:local}
\KwInput {$\mathcal{P}$, $\zeta$}
    $\mathcal{F}$ $\gets$ $\textbf{DetectFrontiers}$($\mathcal{P},\ \mathbb{M_{L}}$) \\
    $\mathbb{M_{G}}$ $\gets$ $\textbf{RegisterGlobalFrontiers}$($\mathcal{F}$) \\
    \eIf{$\mathcal{F} \neq \varnothing$}{
    
        $P_{v}P_{s}$ $\gets$ $\textbf{TunePlanningSpace}$ ($\zeta$, $\mathcal{P}$) \\
        $\mathcal{N}$ $\gets$ $\textbf{SampleNodes}$ ($P_{v}P_{s}$) \\    
        $\mathcal{E}$ $\gets$ $\textbf{FormConnections}$ ($\mathcal{N},\ \mathbb{M_{L}}$) \\
        $\mathbb{S_{G}}^{i}$ $\gets$ $\textbf{BuildSubGraph}$ ($\mathcal{N},\ \mathcal{E}$) \\
        
        \For{node $n$ : $\mathbb{S_{G}}^{i}$} {
            $\textbf{NearestNeighboursLookup}$ ($\mathcal{F}$) \\ 
            \If{neighbouring\_frontiers > $k^{nn}_{threashold}$}{
                $n$ $\gets$ is\_frontier \\
                $n$ $\gets$ $\textbf{ComputeVolumetricGain}$ ($\mathbb{M_{L}}$) 
            }
        }
        $P_{L}$ $\gets$ $\textbf{DijkstraPathsToFrontierNodes}$ ($\mathbb{S_{G}}$) \\
        $\lambda_{L}$ $\gets$ $\textbf{ComputeCandidatePath}$ ($P_{L}$) \\
        $\lambda_{L}^{k}$ $\gets$ $\textbf{SegmentPath}$ ($\lambda_{L}$) \\
       $\lambda^{k}_{refined}$ $\gets$ $\textbf{RefinePathSegment}$ ($\mathbb{ST}_{image}$) \\
        nMPC $\gets$ $\textbf{ExecutePathSegment}$ ($\lambda^{k}_{refined}$) \\
        $\mathbb{G_{G}}$ $\textbf{UpdateGlobalGraph}$ ($\mathbb{S_{G}}^{i}$)
    }
    {
        $\textbf{RunGlobalRepositioningPlanner}$
    }
\normalsize
\end{algorithm} 
\subsection{Global Re-Positioning Planning} \label{global}

\begin{figure}
    \centering
        \includegraphics[width=\linewidth]{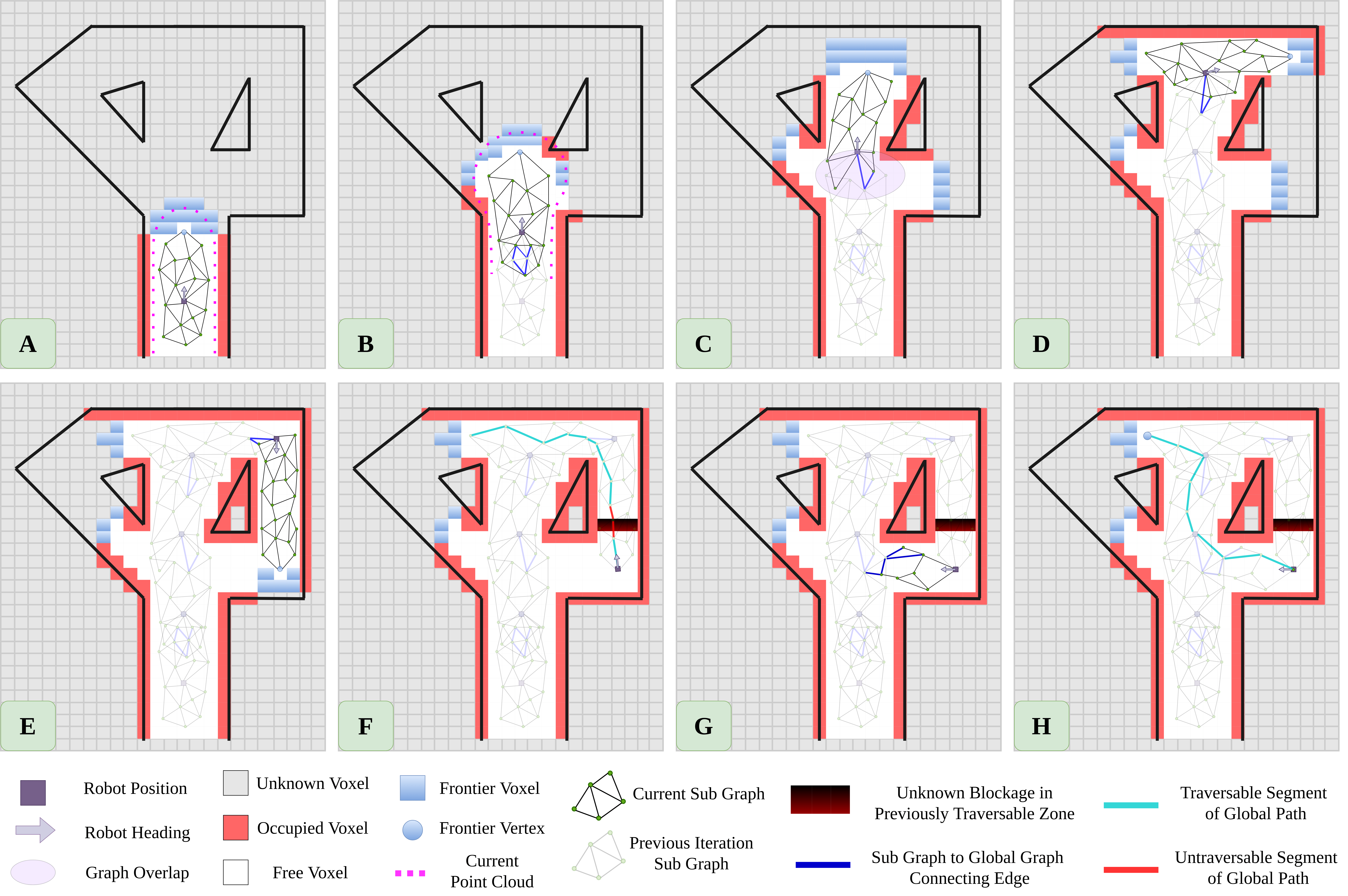}
    \caption{STAGE planner overview visualization in \textbf{A-H}. Steps \textbf{A-E} depict the sequential global graph build through the association of the local sub-graphs. Step \textbf{F} visualizes the concept of obstructed path in the global graph during global re-positioning and the removal of connected nodes of the global graph in the untraversable area. Step \textbf{G} visualizes the sub-graph on an unexplored nearby alternative area to reach the re-positioning global frontier. Step \textbf{H} depicts the global path-way segments on the adapted global graph.}
    \label{fig:concept_graph}
    \vspace{-1.5em}
\end{figure}

\begin{algorithm}
\small
\SetAlgoLined
\caption{Global Re-positioning} \label{algo:global}
\KwInput {$\mathbb{S_{G}}^{i}$, $\mathbb{S_{G}}^{i-1}$, $\zeta$}
    $\mathcal{F}$ $\gets$ $\textbf{DetectFrontiers}$ ($\mathcal{P},\ \mathbb{M_{L}}$) \\
    \eIf{$\mathcal{F} \neq \varnothing$}
    {
        $\mathbb{M_{G}}$ $\gets$ $\textbf{RegisterGlobalFrontiers}$ ($\mathcal{F}$) \\
        \If{\FuncSty{UpdateGlobalGraph}}
        { 
            $\mathbb{G_{G}}$ $\gets$ $\textbf{InsertSubGraphs}$ ($\mathbb{S_{G}}^{i}$, $\mathbb{S_{G}}^{i-1}$) \\
            $\mathcal{N}_{overlap}$ $\gets$ $\textbf{DetectOverlap}$ ($\mathbb{S_{G}}^{i}$, $\mathbb{S_{G}}^{i-1}$) \\
            $\mathbb{G_{G}}$ $\gets$ $\textbf{FormConnections}$ ($\mathcal{N}_{overlap}$) \\
        }
        $P_{G}$ $\gets$ $\textbf{DijkstraPathsToFrontierNodes}$ ($\mathbb{G_{G}}$) \\
        $\lambda_{G}$ $\gets$ $\textbf{EvaluateGlobalPathways}$ ($P_{G}$) \\
        $\lambda_{G}^{k}$ $\gets$ $\textbf{SegmentPath}$ ($\lambda_{G}$) \\
        $\textbf{EvaluateSegmentTraversability}$ ($\mathbb{M_{L}}$, $\lambda_{G}^{k}$) \\
        \If{$\lambda_{G}^{k}$\_is\_UNTRAVERSABLE}
        {
            $O_{ss}$ $\gets$ $\textbf{OrientedSamplingSpace}$ ($\zeta$, $\lambda_{G}^{k}$) \\
            $\textbf{SampleNodes}$ ($O_{ss}$) \\
            $\textbf{FindAlternativeSegment}$ () \\
            \If{!alt\_segment\_traversable}{
                $P_{G}$ $\gets$ $\textbf{RemovePathWay}$ ($\lambda_{G}$) \\
                $\mathbb{G_{G}}$ $\gets$ $\textbf{UpdateGraph}$ ($P_{G}$) \\
                $\lambda_{G}$ $\gets$ $\textbf{EvaluateGlobalPathways}$ ($P_{G}$) \\
                $\lambda_{G}$ $\gets$ $\textbf{RePlan}$ ($\mathbb{G_{G}}$) \\
            }
            $\lambda^{k}_{refined}$ $\gets$ $\textbf{RefinePathSegment}$ ($\mathbb{ST}_{image}$) \\
            $\textbf{EvaluateSegmentTraversability}$ ($\mathbb{M_{L}}$, $\lambda_{G}^{k}$) \\
            nMPC $\gets$ $\textbf{ExecutePathSegment}$ ($\lambda^{k}_{refined}$) \\  
        }
    }
    {
        $\textbf{PlanReturnToHome}$ () 
    }
\normalsize
\end{algorithm} 

\begin{figure*}[htbp!]
    \centering
        \includegraphics[width=\textwidth]{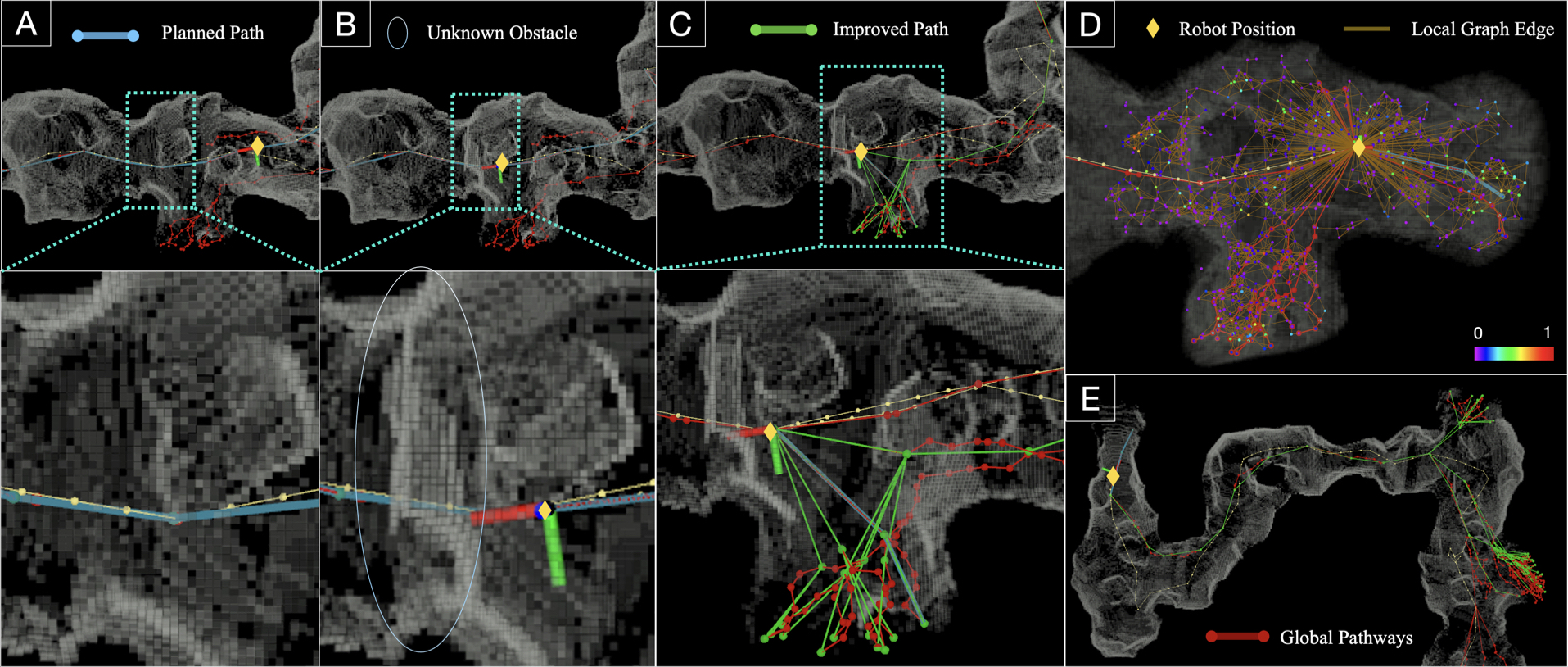}
    \caption{STAGE planner's adaptive re-planning behaviour in presence of unknown change to the global map. The obstacle in [\textbf{B}] is added when robot was out of line of sight. In [\textbf{C}] re-planned path ways are highlighted in green.}
    \vspace{-1.5em}
    \label{fig:sims_collage}
    
\end{figure*}
As mentioned in \autoref{subsec:local_exploration} the global re-positioning to potential frontier is triggered when local exploration is insufficient.  For global navigation the novelty of the STAGE planner stems from the re-utilization of previously built sub-graphs $\mathbb{S_{G}}^{i}$ \& $\mathbb{S_{G}}^{i-1}$ to build a dense global graph $\mathbb{G_{G}}$. To connect the sub-graphs, we first detect the overlap in tuned planning spaces $P_{v}P_{s}$ corresponding to the latest and previous iteration sub-graphs. As shown in \autoref{algo:global} the nodes belonging to the overlapping region are connected with the neighbours from the previous sub-graph to form continuous connection in $\mathbb{G_{G}}$ from robot's current position to the location of the mission base station. A visual representation of the graph overlap and global graph building step is shown in Fig. \ref{fig:concept_graph}. Through planning steps \textbf{A} to \textbf{E} in Fig. \ref{fig:concept_graph}, the robot builds $\mathbb{G_{G}}$ by re-utilizing $\mathbb{S_{G}}$. Through Dijkstra shortest paths algorithm, we further calculate path ways leading to all potential global frontiers from robot's current state. The reward $\mathbb{R}$ is computed for each node along the global path way $\lambda_{G} \in P_{G}$ to compute global re-position reward for $\lambda_{\mathbb{R}}$. As described in (\ref{eqn:global_repos_reward}) the re-position reward is further penalized through path length $d_{\lambda}$ where, $W_{d}$ is the weight associated with path length penalty.  
\begin{align} \label{eqn:global_repos_reward}
    \lambda_{\mathbb{R}} = \sum_{p=1}^{p} \mathbb{R}_{p} - W_{d} * d_{\lambda}
\end{align}
\subsubsection{Uncertainty Aware Re-Planning} \label{subsubsec:re_planning}
As motivated above, the STAGE planner aims to handle dynamic changes in the explored map with the merit to fulfil global re-positioning without utilizing full $\mathbb{M_{G}}$ during path re-planning. Dynamic changes of the scene can occur in such types of environments including rockfalls, blocked passages due to machine breakdowns, doors and other types of pathway obstructions. As it might happen during a change of the scene, the section of the global map $\mathbb{M_{\delta}} \subset \mathbb{M_{G}}$ is changed after the robot had initially mapped the $\mathbb{M_{\delta}}$. Since previously it had traversed through $\mathbb{M_{\delta}}$, the robot had built the global graph where the global path ways $P_{G}$ pass through $\mathbb{M_{\delta}}$. If an unknown uncertainty $\Upsilon_{u}$ is introduced in the $\mathbb{M_{\delta}}$ at a time instance that the robot is not present in the scene, the planned global pathways $P_{G}$ will be in collision with the unknown $\Upsilon_{u}$. If the robot blindly follows a path way $\lambda_{G} \in P_{G}$, it will result in collision with the unknown change in the global map as shown in planning step \textbf{F} of Fig. \ref{fig:concept_graph}.

To address this, we introduce adaptive graph updates without utilizing full $\mathbb{M_{G}}$. As depicted in \autoref{algo:global}, before the path segment execution, the planner constructs an Oriented Sampling Space $O_{ss}$ oriented from $\zeta$ towards the end point of current path segment $\lambda^{k}$. A ray casting operation is performed from $\zeta$ along $\lambda^{k}$. If this ray casting reports collision with occupied or unknown voxel, intuitively it is assumed that the $\mathbb{M_{\delta}}$ is changed. Since the current segment $\lambda^{k}$ is now untraversable, we sample new nodes in $O_{ss}$ to find if an alternative path segment can be generated to plan the path around $\Upsilon_{u}$. If no such segment is found, then the path way is marked untraversable as $\lambda_{ut}$ in the $P_{G}$ followed by global graph update $\mathbb{G_{G}}$ that removes the edges that form $\lambda_{ut}$. The planner re-evaluates the updated global graph to compute new $\lambda_{G}$ to continue exploration while registering the unknown change $\Upsilon_{u}$ in the global graph to further discard edges in newly discovered untraversable zone. A visual representation of this re-planning process is shown in the planning steps \textbf{F}, \textbf{G} and \textbf{H} of Fig. \ref{fig:concept_graph}.



\begin{figure*}[t]
    \centering
        \includegraphics[width=\textwidth]{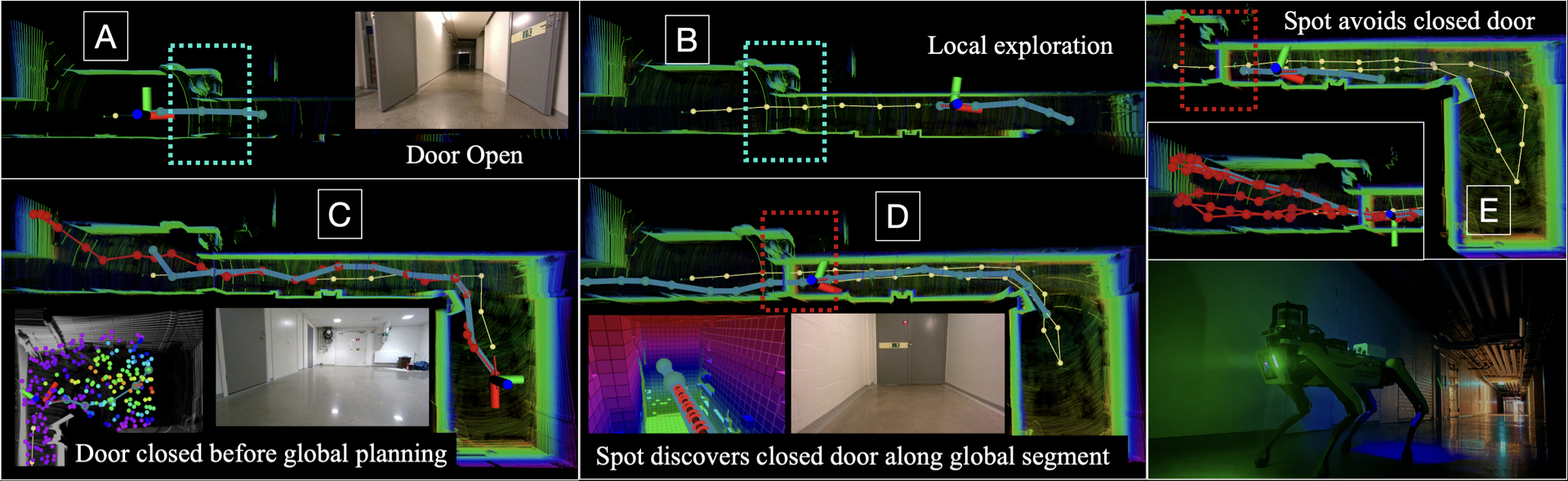}
    \caption{Experimental evaluation of STAGE planner with Boston Dynamics spot robot.}
    \label{fig:experiment_collage}
    \vspace{-1.5em}
\end{figure*}


\section{STAGE Evaluation}

The STAGE planner was evaluated in simulations and real life experiment to demonstrate A) Scalable efficient exploration in large SubT settings and B) Adaptive global re-positioning in the presence of unknown change in the explored map. The conducted simulations and experimental evaluation of stage planner can be watched at \url{https://youtu.be/NC4r_hhfZ-M?si=cFwbdwqdLFaRnil_}. STAGE planner is also platform agnostic, evaluated for both aerial robot in simulation and legged robot in real life experiments. The simulations were carried out with in gazebo simulator using the RotorS\cite{furrer2016rotors} Robot Operating System (ROS) package. The robot is equipped with a Velodyne VLP16 3D LiDAR to generate point cloud as an input for STAGE planner. The goal of the simulation evaluation was to have the robot explore unknown subterranean cave environment that has multiple junctions and varying topology like narrow passages and large void-like structures. To demonstrate the adaptive global re-planning in presence of uncertainty in the explored map, we block the passage way through which the robot has planned global path. As shown in Fig. \ref{fig:sims_collage} [\textbf{A}] the robot has planned path in traversable space of the global map $\mathbb{M_{G}}$. As the robot executes the planned global path $\lambda_{G}$, we purposefully block the passage through which the robot has already planned $\lambda_{G}$. Normally in global re-positioning maneuver, the robots blindly follow the global or homing paths. As mentioned in \autoref{sec:methodology} in contrast to the state-of-the-art approaches, the proposed planner segments the global path and evaluates each segment before execution to handle any uncertainty present. As shown in the Fig. \ref{fig:sims_collage} [\textbf{B}], the proposed method evaluates the immediate segment within point cloud visibility and discovers that a change $\Upsilon_{u}$ in the map has occurred. Then, the next step is to check if alternate paths can be found to reach the goal of $\lambda_{G}$. Since the passage way was completely blocked, the robot takes into consideration $\Upsilon_{u}$ and re-plans global path way to another global frontier in the vicinity. As highlighted in the Fig. \ref{fig:sims_collage} [\textbf{C}], the robot finds alternate improved pathways that account for newly discovered change $\Upsilon_{u}$. As a result, the robot is able to safely globally re-position and continue exploration. A highlight of the information gain associated with frontier nodes of sub graph is shown in  Fig. \ref{fig:sims_collage} [\textbf{D}]. The information gain scales from 0 to 1 and higher informative path in free space is planned as shown in Fig. \ref{fig:sims_collage} [\textbf{D}]. Finally in \textbf{E} we show instance of exploration where the robot is locally exploring and has candidate paths to global exploration are shown in green. 
\begin{figure}[h!]
    \centering
    \subfigure
    {
        \includegraphics[width=0.465\linewidth]{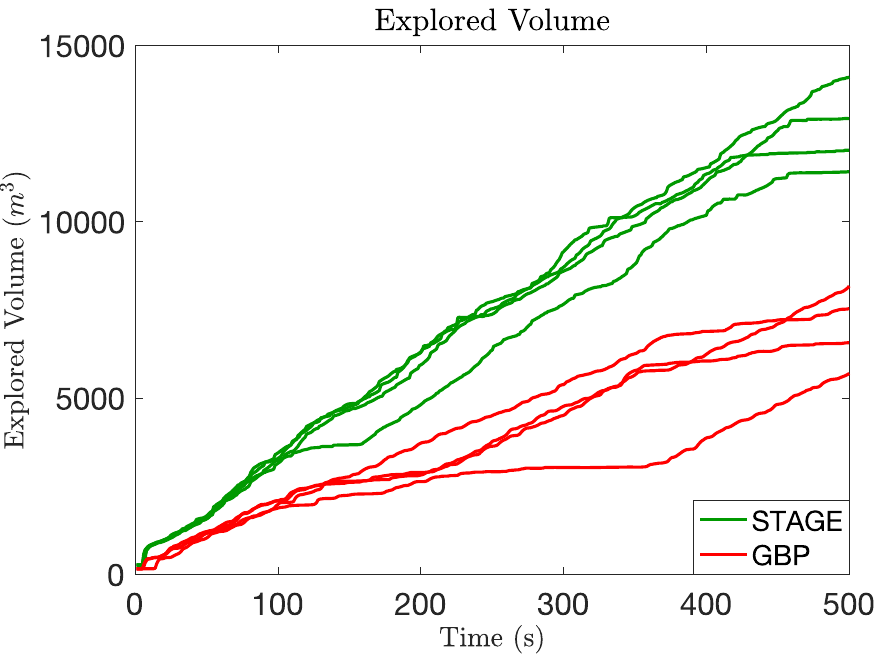}
    }
    \subfigure
    {
        \includegraphics[width=0.465\linewidth]{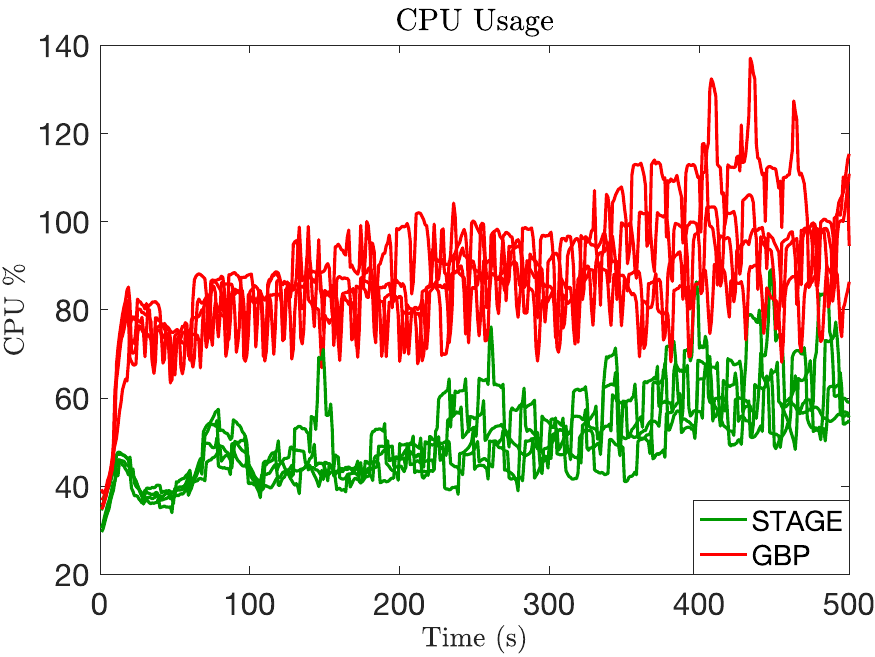}
    }
    \caption{Explored volume and Resource utilization comparison.}
    \vspace{-1.5em}
    \label{fig:explorationmetric}
\end{figure}
The experimental evaluation of the STAGE planner was performed in the underground corridors of university building, located at the Lule{\aa} University of Technology campus in Sweden. The hardware setup for the field tests used the Boston Dynamics Spot robot, configured with our own sensor and computation kit on top. The Spot kit also uses an Intel NUC 10 BXNUC10I5FNKPA2 for computation, a Velodyne Puck Lite 3d LiDAR, a RealSense D455 mounted forward and a VectorNav VN-100 IMU. To highlight the novelty of the STAGE planner we choose a scenario where the Spot robot locally explores a long corridor that is connected to open room at the end (replicating the void like topology) and while the robot is out of sight of the corridor, we purposefully block the passage by closing the door. In the Fig. \ref{fig:experiment_collage} [\textbf{A}], when the Spot robot starts the exploration the door is open (area highlighted in blue box). As the mission progresses Fig. \ref{fig:experiment_collage} [\textbf{B}] the Spot robot explores the corridor and reaches the end where open room is found. While the robot is out of line of sight as shown in Fig. \ref{fig:experiment_collage} [\textbf{C}], the door is purposefully closed (area highlighted in blue box). At this instance, the Spot robot has planned a global re-positioning path to continue exploration in another corridor with the previous information that the passage is traversable. As the spot robot navigates along the global path $\lambda_{G}$, in Fig. \ref{fig:experiment_collage} [\textbf{D}] it discovers an untraversable segment. No alternate paths could be found as the door was fully closed. Therefore, as shown in Fig. \ref{fig:experiment_collage} [\textbf{E}] the Spot robot re-positions itself to explore residual frontier nodes in the previous corridor before completing the exploration mission.
Since the STAGE planner is designed to be scalable and platform agnostic, we have compared the performance of in exploration and resource utilization with the State of the Art graph based planner [GBP] \cite{dang2019graph}. We perform multiple runs of the STAGE planner and GBP to explore a subterranean cave environment with equal mapping and robot dynamics and control parameters. The results of the performance comparison are highlighted through Fig. \ref{fig:explorationmetric}. Since GBP is also a graph based planner and has global planning functionality, it explores considerable portion of the environment. However, due to STAGE planner's direct point cloud visibility based rapid local exploration and efficient global re-positioning, STAGE planner shows comparatively higher exploration gain. Apart from the standard exploration volume gain metric, we also compare the CPU utilization of the two algorithms in the same runs. Since STAGE planner adapts a novel sub-graph overlap based global graph generation and re-planning, the STAGE planner consumes considerably less CPU to run both \autoref{algo:local} and \autoref{algo:global} in the same ROS node.



\section{Conclusion}

Through this article we propose the STAGE planner, a scalable novel navigation scheme for large scale exploration in environments with dynamic scene changes. STAGE is a two layered graph based framework with memory efficient management of the local and global layers. More specifically, the local sub-graphs are adaptively tuned to local bounds based on the detected frontiers on the direct pointcloud visibility, enabling fast traversal towards local frontiers. The global graph is built using minimal node-edge information exchange only on the overlapping regions of sequential sub-graphs. STAGE is also traversability aware planer able to handle scene changes (e.g. blocked pathways) in previously explored areas. It adaptively updates the obstructed part of the global graph from traversable to not-traversable. Each path in the global graph is divided in multiple segments, where each one is collision checked through oriented sample space to remove edges from connected nodes in cases of obstructions. In such cases, the exploration behavior is directing the robot to follow another route in the global re-positioning phase through path-way updates in the global graph. The performance of the method has been evaluated in simulation runs using aerial vehicles compared with another state-of-the-art exploration algorithm, while it has also been deployed in real-world corridor scene using a legged robot.


\bibliographystyle{ieeetr}

\bibliography{References}

\end{document}